%% file: arxiv.tex
\documentclass[10pt,twocolumn,letterpaper]{article}

\usepackage{times}
\usepackage{epsfig}
\usepackage{graphicx}
\usepackage{amsmath}
\usepackage{amssymb}
\usepackage{array}
\usepackage{multirow}
\usepackage{url}
\usepackage{wrapfig}
\usepackage{booktabs}
\usepackage{subfig}
\usepackage[pagebackref=true,breaklinks=true,letterpaper=true,colorlinks,bookmarks=false]{hyperref}

\begin{document}

\title{3D Semantic Segmentation with\\ Submanifold Sparse Convolutional Networks}

\author{
  Benjamin Graham\thanks{Facebook AI Research} ~,
  Martin Engelcke\thanks{University of Oxford, intern at Facebook AI Research} ~,
  Laurens van der Maaten\footnotemark[1]
  \\
  \tt\small{\{benjamingraham, lvdmaaten\}@fb.com} \\
  \tt\small{martin@robots.ox.ac.uk}\\
}
\maketitle

\begin{abstract}
\input{abstract}
\end{abstract}

\input{intro}
\input{related}
\input{problem_statement}
\input{ssc}
\input{semantic}
\input{largescale}
\input{nyu}
\input{conclusions}

\small{

}
\end{document}

%% file: abstract.tex
Convolutional networks are the de-facto standard for analyzing spatio-temporal data such as images, videos, and 3D shapes. Whilst some of this data is naturally dense (e.g., photos), many other data sources are inherently sparse. Examples include 3D point clouds that were obtained using a LiDAR scanner or RGB-D camera. Standard ``dense'' implementations of convolutional networks are very inefficient when applied on such sparse data. We introduce new sparse convolutional operations that are designed to process spatially-sparse data more efficiently, and use them to develop spatially-sparse convolutional networks. We demonstrate the strong performance of the resulting models, called submanifold sparse convolutional networks (SSCNs), on two tasks involving semantic segmentation of 3D point clouds.
In particular, our models outperform all prior state-of-the-art on the test set of a recent semantic segmentation competition.

%% file: intro.tex
\section{Introduction}

Convolutional networks (ConvNets) constitute the state-of-the art method for a wide range of tasks that involve the analysis
of data with spatial and/or temporal structure, such as photos, videos,
or 3D surface models. While such data frequently comprises a densely populated (2D or 3D) grid, other datasets are naturally sparse. For instance, handwriting is made up of one-dimensional lines in two-dimensional space, pictures made by RGB-D cameras are three-dimensional point clouds, and polygonal mesh models
form two-dimensional surfaces in 3D space.

The curse of dimensionality applies, in particular,
to data that lives on grids that have three or more dimensions: the number of points on the grid grows exponentially with its dimensionality. In such scenarios, it becomes increasingly important to exploit data sparsity whenever possible in order to reduce the computational resources needed for data processing. Indeed, exploiting sparsity is paramount when analyzing, \emph{e.g.}, RGB-D videos which are sparsely populated 4D structures.

\begin{figure}[t!]
    \centering
    \subfloat{{\includegraphics[width=0.33\columnwidth,trim={3cm 3cm 3cm 1cm},clip]{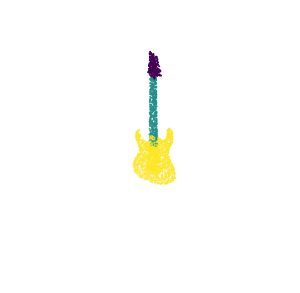}}}
    \subfloat{{\includegraphics[width=0.33\columnwidth,trim={3cm 3cm 3cm 2cm},clip]{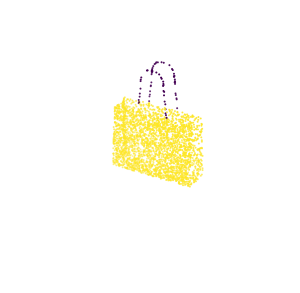}}}
    \subfloat{{\includegraphics[width=0.33\columnwidth,trim={3cm 3cm 3cm 2cm},clip]{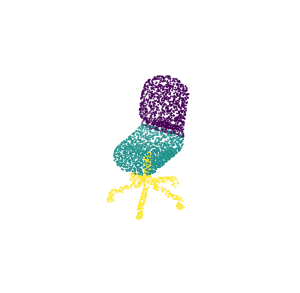}}}
    \\
    \subfloat{{\includegraphics[width=0.33\columnwidth,trim={3cm 3cm 3cm 3cm},clip]{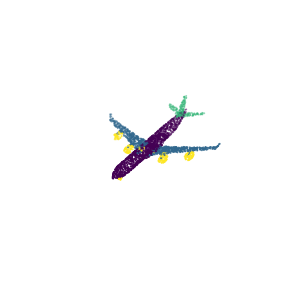}}}
    \subfloat{{\includegraphics[width=0.33\columnwidth,trim={3cm 3cm 3cm 3cm},clip]{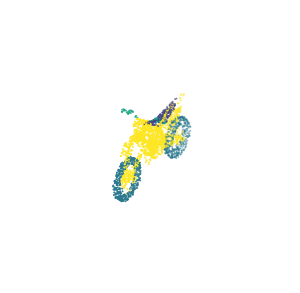}}}
    \subfloat{{\includegraphics[width=0.33\columnwidth,trim={3cm 3cm 3cm 3cm},clip]{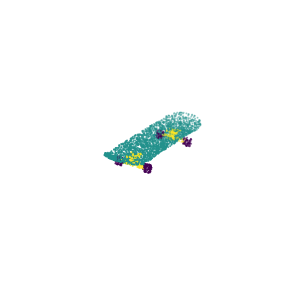}}}
    \caption{Examples of 3D point clouds of objects from the ShapeNet part-segmentation challenge~\cite{yi2017large}. The colors of the points represent the part labels.}
    \label{fig:data}
\end{figure}

Traditional convolutional network implementations are optimized for data that lives on densely populated grids, and cannot process
sparse data efficiently. More recently, a number of convolutional network implementations have been presented that are tailored to work efficiently on sparse data \cite{graham2015sparse,riegler2016octnet,engelcke2017vote3deep}.
Mathematically, some of these implementations are identical to regular convolutional networks, but they require fewer computational resources in terms of FLOPs and/or memory \cite{graham2015sparse,engelcke2017vote3deep}. Prior work uses a sparse version of the \texttt{im2col} operation that restricts computation and storage to ``active'' sites \cite{graham2015sparse}, or uses the voting algorithm from \cite{wang2015voting} to prune unnecessary multiplications by zeros \cite{engelcke2017vote3deep}. OctNets \cite{riegler2016octnet} modify the convolution operator to produce ``averaged'' hidden states in parts of the grid that are outside the region of interest.

One of the downsides of prior sparse implementations of convolutional networks is that they ``dilate'' the sparse data in every layer by applying ``full'' convolutions. In this work, we show that it is possible to create convolutional networks that
\emph{keep the same level of sparsity throughout the network}.
To this end, we develop a new implementation for performing \emph{sparse convolutions} (SCs) and introduce a novel convolution operator termed \emph{submanifold sparse convolution} (SSC).\footnote{These operators appeared earlier in an unpublished technical report \cite{graham2017submanifold}.}
We use these operators as the basis for submanifold sparse convolutional networks (SSCNs) that are optimized for efficient semantic segmentation of 3D point clouds, \emph{e.g.}, on the examples shown in Figure~\ref{fig:data}.

In Table~\ref{tab:seg_results}, we present the performance of SSCNs on the test set of a recent part-based segmentation competition~\cite{yi2017large} and compare it to some of the top-performing entries in the competition: SSCNs outperform all of these entries. Source code for our library is publicly available online\footnote{\url{https://github.com/facebookresearch/SparseConvNet}}.

\begin{table}[t]
  \centering
  \small
  \begin{tabular}{lc}
    \toprule
    {\bf Method} & {\bf Average IoU}\\
    \midrule
    NN matching with Chamfer distance & 77.57\% \\
    Synchronized Spectral CNN \cite{yi2016syncspeccnn} & 84.74\% \\
    \midrule
    Pd-Network (extension of Kd-Network \cite{klokov2017escape}) & 85.49\% \\
    Densely Connected PointNet (extension of \cite{qi2016pointnet}) & 84.32\% \\
    PointCNN & 82.29\% \\
    \midrule
    Submanifold SparseConvNet (Section \ref{subsec:aligned}) & \textbf{85.98\%} \\
    \bottomrule
  \end{tabular}
    \caption{Average intersection-over-union (IoU) of six approaches on the test set of a recent part-based segmentation competition on ShapeNet~\cite{yi2017large}. Higher is better. Our SSCNs outperform all alternative approaches.}
  \label{tab:seg_results}
\end{table}

%% file: related.tex
\section{Related Work}

Our work primarily builds upon previous literature on sparse convolutional networks \cite{engelcke2017vote3deep, graham2015sparse}, and image segmentation using dense convolutional networks \cite{long2015fully, ronneberger2015unet, yu2015multi}.
Examples of applications of dense 3D convolutions on volumetric data include classification \cite{maturana2015voxnet} and segmentation \cite{cciccek20163d}; these methods suffer from high memory usage and slow inference, limiting the size of models that can be used.

Methods for processing 3D point clouds without voxelization have also been developed \cite{klokov2017escape,qi2016pointnet}. This may seem surprising given the dominance of ConvNets for processing 2D inputs; it is likely due to the computational obstacles involved in using dense 3D convolutional networks.

Prior work on sparse convolutions implements a convolutional operator that increases the number of active sites with each layer~\cite{engelcke2017vote3deep,graham2015sparse}. In~\cite{graham2015sparse}, all sites that have at least one ``active'' input site are considered as active. In \cite{engelcke2017vote3deep}, a greater degree of sparsity is attained {\em after} the convolution has been calculated by using ReLUs and a special loss function.
In contrast, we introduce \emph{submanifold} sparse convolutions that fix the location of active sites so that the sparsity remains unchanged for many layers. We show that this makes it practical to train deep and efficient networks similar to VGG networks~\cite{journals/corr/SimonyanZ14a} or ResNets~\cite{he2016deep}, and that it is well suited for the task of point-wise semantic segmentation.

OctNets~\cite{riegler2016octnet} are an alternative form of sparse convolution. Sparse voxels are stored in oct-trees: a data structure in which the grid cube is progressively subdivided into
$2^3$ smaller sub-cubes until the sub-cubes are either empty or contain a single active site.
OctNet operates on the surfaces of empty regions, so a size-$3$ OctNet convolution on an empty cube of size $8^3$ requires 23\% of the calculation of a dense 3D convolution. Conversely, submanifold convolutions require no calculations in empty regions.

Another approach to segmenting point clouds is to avoid voxelizing the input, which may lead to a loss of information due to the finite resolution. This can be done by either using carefully selected data structures such as Kd-trees~\cite{klokov2017escape}, or by directly operating on the unordered set of points \cite{qi2016pointnet}.
Kd-Networks~\cite{klokov2017escape} build a Kd-tree by recursively partitioning the space along the axis of largest variation until each leaf of the tree represents one input point. This takes O$(N\log N)$ time for $N$ input points.
PointNet~\cite{qi2016pointnet} uses a pooling operation to produce a global feature vector.

Fully convolutional networks (FCNs) were proposed in \cite{long2015fully} as a method of 2D image segmentation; FCNs make use of information at multiple scales to preserve low-level information to accurately delineate object boundaries.
U-Nets \cite{ronneberger2015unet} extend FCNs by using convolutions to more accurately merge together the information from the different scales before the final classification stage; see Figure~\ref{fig:architecture_diagram}.

%% file: problem_statement.tex
\section{Spatial sparsity for ConvNets}
We define a $d$-dimensional convolutional network as a network that
takes as input a $(d+1)$-dimensional tensor: the input tensor contains $d$ spatio-temporal
dimensions (such as length, width, height, time, \emph{etc.}) and one additional feature-space
dimension (\emph{e.g.}, RGB color channels or surface normal vectors).
The input corresponds to a $d$-dimensional grid of \emph{sites}, each of which is
associated with a feature vector. We define a site in the input to be \emph{active} if any element in the feature vector is not in its \emph{ground state}, \emph{e.g.}, if it is non-zero\footnote{Note that the ground state does not necessarily have to be zero, in particular, when convolutions with a bias term are used.}. In many problems, thresholding
may be used to eliminate input sites at which the feature
vector is within a small distance from the ground state. Note that even though the input tensor is $(d+1)$-dimensional,
activity is a $d$-dimensional phenomenon: entire lines along the feature dimension
are either active or inactive.

Similarly, the hidden layers of a $d$-dimensional convolutional network are represented by $d$-dimensional
grids of feature-space vectors. When propagating the input data through the network,
a site in a hidden layer is active if any of the sites
in the layer that it takes as input is active. (Note that when using size-3 convolutions,
each site is connected to $3^d$ sites in the hidden layer below.) Activity in a hidden layer thus follows an inductive definition in which each layer determines the set of active states in the next layer. In each hidden layer, inactive
sites all have the same feature vector: the one corresponding to the ground state.
The value of the ground state only needs to be calculated once per forward pass at training time, and only once for all forward passes at test time. This allows for substantial savings in computational and memory requirements; the exact savings depend on data sparsity and network depth.

However, we argue that the framework described above is unduly restrictive, in particular, because the convolution
operation has not been modified to accommodate the sparsity
of the input data. If the input data contains a single active site, then
after applying a $3^d$ convolution, there will be $3^{d}$ active
sites. Applying a second convolution of the same size will yield $5^{d}$ active sites,
and so on. This rapid growth of the number of active sites is a poor prospect when implementing
modern convolutional network architectures that comprise tens or even hundreds of convolutional layers, such as VGG networks, ResNets, or DenseNets \cite{ResNet,DenseNet,journals/corr/SimonyanZ14a}.

\begin{figure}
\begin{centering}
\includegraphics[width=0.32\columnwidth]{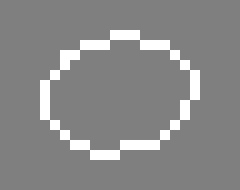}
\includegraphics[width=0.32\columnwidth]{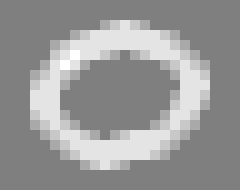}
\includegraphics[width=0.32\columnwidth]{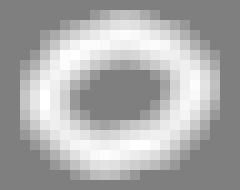}
\caption{Example of ``submanifold'' dilation. \textbf{Left:} Original curve. \textbf{Middle:} Result of applying a regular $3 \times 3$ convolution with weights $1/9$. \textbf{Right:} Result of applying the same convolution again. The example shows that regular convolutions substantially reduce the sparsity of the features with each convolutional layer.}\label{6pics}
\end{centering}
\end{figure}

Of course, convolutional networks are not often applied to inputs that only have a single active site, but the aforementioned dilation problems are equally problematic when the input data comprises one-dimensional
curves in spaces with two or more dimensions, or two-dimensional surfaces in three or more dimensions. We refer to this problem as the ``submanifold dilation problem'', which is illustrated in Figure \ref{6pics}. The figure illustrates that even when we apply small $3\times3$ convolutions on this grid, the sparsity of the grid rapidly disappears.

%% file: ssc.tex
\section{Submanifold Convolutional Networks}
We explore a simple solution to the submanifold dilation problem that restricts the output of the convolution only to the set of
active input points. A potential problem of this approach is that hidden layers in the network may not receive all the information they require to classify the input data: in particular, two neighboring connected components are treated
completely independently. We resolve this problem by using convolutional networks that incorporate
some form of pooling, or use strided convolutions. Such operations are essential in the sparse convolutional networks\footnote{By ``sparse convolutional networks'', we mean networks designed to operate on spatially-sparse input data. We do not mean networks that have sparse parameter matrices \cite{braindamage,liupensky}.} we investigate, as they allow
information to flow between disconnected components in the input. The closer the components are spatially, the fewer strided operations are necessary for the components to ``communicate'' in the intermediate representations.

\subsection{Sparse Convolutional Operations}
We define a sparse convolution SC($m,n,f,s)$ with $m$ input
feature planes, $n$ output feature planes, a filter size of $f$,
and stride $s$.
An SC convolution computes the set of active
sites in the same way as a regular convolution: it looks for the presence of any active sites
in its receptive field of size $f^{d}$. If the input has size $\ell$
then the output will have size $(\ell-f+s)/s$.
Unlike a regular convolution (and the sparse convolution from \cite{graham2015sparse}), an SC convolution discards the ground state for non-active sites by assuming that the input from those sites
is zero. This seemingly small change to the convolution brings
computational benefits in practice.

\begin{figure}[t!]
    \centering
\includegraphics[width=0.32\columnwidth]{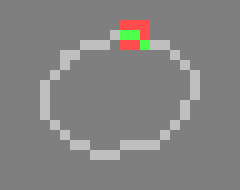}
\includegraphics[width=0.32\columnwidth]{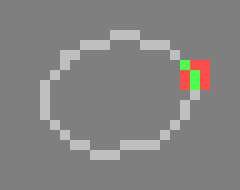}
\includegraphics[width=0.32\columnwidth]{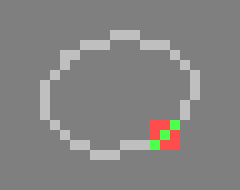}
\caption{SSC$(\cdot,\cdot,3)$ receptive field centered at different active spatial locations. Active locations in the field are shown in green. Red locations are ignored by SSC so the pattern of active locations remains unchanged.}\label{fig:ssc}
\end{figure}

\paragraph{Submanifold sparse convolution.} Next, we define a second type of sparse convolution, which forms
the main contribution of this paper. Let $f$ denote an odd number.
We define a submanifold
sparse convolution SSC$(m,n,f)$ as a modified SC$(m,n,f,s=1)$
convolution. First, we pad the input with $(f-1)/2$ zeros on each side,
so that the output will have the same size as the input. Next, we restrict an output
site to be active if and only if the site at the corresponding
site in the input is active (\emph{i.e.}, if the central site in the receptive field is active). Whenever an output site is determined to be active, its output feature vector is computed by the SSC convolution; see Figure~\ref{fig:ssc} for an illustration. Table \ref{tbl:flops} shows the computational and memory requirements of a regular convolution (C) and our SC and SSC convolutions.

Submanifold sparse convolutions are similar to OctNets \cite{riegler2016octnet} in that they preserve the sparsity structure. However, unlike OctNets, empty space imposes no computational or memory overhead in the implementation of submanifold sparse convolutions.

\paragraph{Other operators.} To construct convolutional networks using SC and SSC, we also need activation functions, batch normalization, and pooling. Activation functions are defined as usual, but are restricted to the set of active sites. Similarly, we define batch normalization in terms of regular batch normalization applied over the set of active sites. Max-pooling MP$(f,s)$ and average-pooling AP$(f,s)$ operations are defined as a variant of SC$(\cdot,\cdot,f,s)$. MP takes the maximum of the zero
vector and the input feature vectors in the receptive field. AP calculates $f^{-d}$ times the sum of the active input vectors. We also define a deconvolution \cite{zeiler10} operation DC$(\cdot,\cdot,f,s)$ as an inverse of the SC$(\cdot,\cdot,f,s)$ convolution. The set of active output sites from a DC convolution is exactly the same as the set of input active sites to the corresponding SC convolution: the connections between input and output sites are simply inverted.

\begin{table}
\centering{}%
\begin{tabular}{ll|ccc}\toprule
 \textbf{Active} & \textbf{Type} & \textbf{C} & \textbf{SC} & \textbf{SSC}\\
\midrule
\multirow{2}{*}{Yes} &  FLOPs & $3^{d}mn$ & $amn$ & $amn$\\
                           &  Memory & $n$ & $n$ & $n$ \\
\midrule
\multirow{2}{*}{No, $a>0$} &  FLOPs & $3^{d}mn$ & $amn$ & 0\\
                           &  Memory & $n$ & $n$ & 0\\
\midrule
\multirow{2}{*}{No, $a=0$} &  FLOPs & $3^{d}mn$ & 0 & 0\\
                           &  Memory & $n$ & 0 & 0\\
\bottomrule
\end{tabular}
\caption{Computational and memory requirements of three convolutional operations: regular convolution (C), sparse convolution (SC), and submanifold sparse convolution (SSC). We consider convolutions with size $f=3$ and padding $s=1$ at a single location
in $d$ dimensions. Notation: $a$ is the number of active inputs to
the spatial location, $m$ the number of input feature planes, and
$n$ the number of output feature planes.}\label{tbl:flops}
\end{table}

\subsection{Implementation}

To implement (S)SC convolutions efficiently, we store the
state of a input/hidden layer in two parts: a hash table\footnote{\url{https://github.com/sparsehash/sparsehash}} and a matrix.
The matrix has size $a \times m$ and contains one row for each of the $a$
active sites. The hash table contains (location, row) pairs
for all active sites: the location is a tuple of integer coordinates,
and the row number indicates the corresponding row in the feature matrix.
Given a convolution with filter size $f$, let $F=\{0,1,\dots,f-1\}^d$ denote the spatial size of the convolutional filter.
Define a \emph{rule book} to be a collection $R=(R_{i}:i\in F)$ of $f^d$ integer matrices each with two columns.
To implement an SC($m,n,f,s)$ convolution, we:

\begin{enumerate}
\item Iterate once through the input hash-table. We build the output hash table and rule book on-the-fly by iterating over
points in the input layers, and all the points in the output layer that can see them.
When an output site is visited for the first time, a new entry is created in the output hash table.
For each active input point $x$ located at point $i$ in the receptive field of an output point $y$, add a row
(input-hash$(x)$, output-hash$(y)$) to element $R_i$ of the rule book.

\item Initialize the output matrix to all zeros. For each $i\in F$, there is a parameter matrix $W^{i}$ with
size $m\times n$. For each row $(j,k)$ in $R_i$, multiply the
$j$-th row of the input feature matrix by $W^{i}$ and add
it to the $k$-th row of the the output feature matrix. This
can be implemented very efficiently on GPUs because it is a
matrix-matrix multiply-add operation.
\end{enumerate}

To implement an SSC convolution, we re-use the input hash table for the
output, and construct an appropriate rule book. Note that because the sparsity pattern does not change, the same rule book
can be re-used in VGG/ResNet/DenseNet networks until a pooling or subsampling layer is encountered in the architecture.

If there are $a$ active points in the input layer, the cost
of building the input hash-table is $O(a)$. For FCN and U-Net
networks, assuming the number of active sites reduces by a multiplicative
factor with each downsampling operation, the cost of building all the
hash-tables and rule-books is also $O(a)$ regardless of the depth of
the network.

The above implementation differs from \cite{graham2015sparse} in that the cost of calculating an output site is proportional to the number of active inputs, rather than the size of the receptive field. For SC convolutions this is similar to the voting algorithm \cite{wang2015voting,engelcke2017vote3deep} - the filter weights are never multiplied with inactive input locations - but for SSC convolutions, this implementation is less computationally intensive than voting as there is no interaction between active input locations and inactive neighboring output locations.

%% file: semantic.tex
\section{Submanifold FCNs and U-Nets\\for Semantic Segmentation}\label{sec:semantic}

Three-dimensional semantic segmentation involves the segmentation of 3D objects or scenes represented as point clouds into their constituent parts; each point in the input cloud must be assigned a part label.
As substantial progress has been made in the segmentation of 2D images using convolutional neural networks~\cite{long2015fully, ronneberger2015unet, yu2015multi}, interest in the problem of 3D semantic segmentation has grown recently. This interest was fueled, in particular, by the release of a new dataset for the part-based segmentation of 3D objects, and an associated competition~\cite{yi2017large}.

\begin{figure*}[t]
    \centering
    \subfloat[Submanifold sparse FCN.]{{\includegraphics[width=0.35\textwidth,height=4cm]{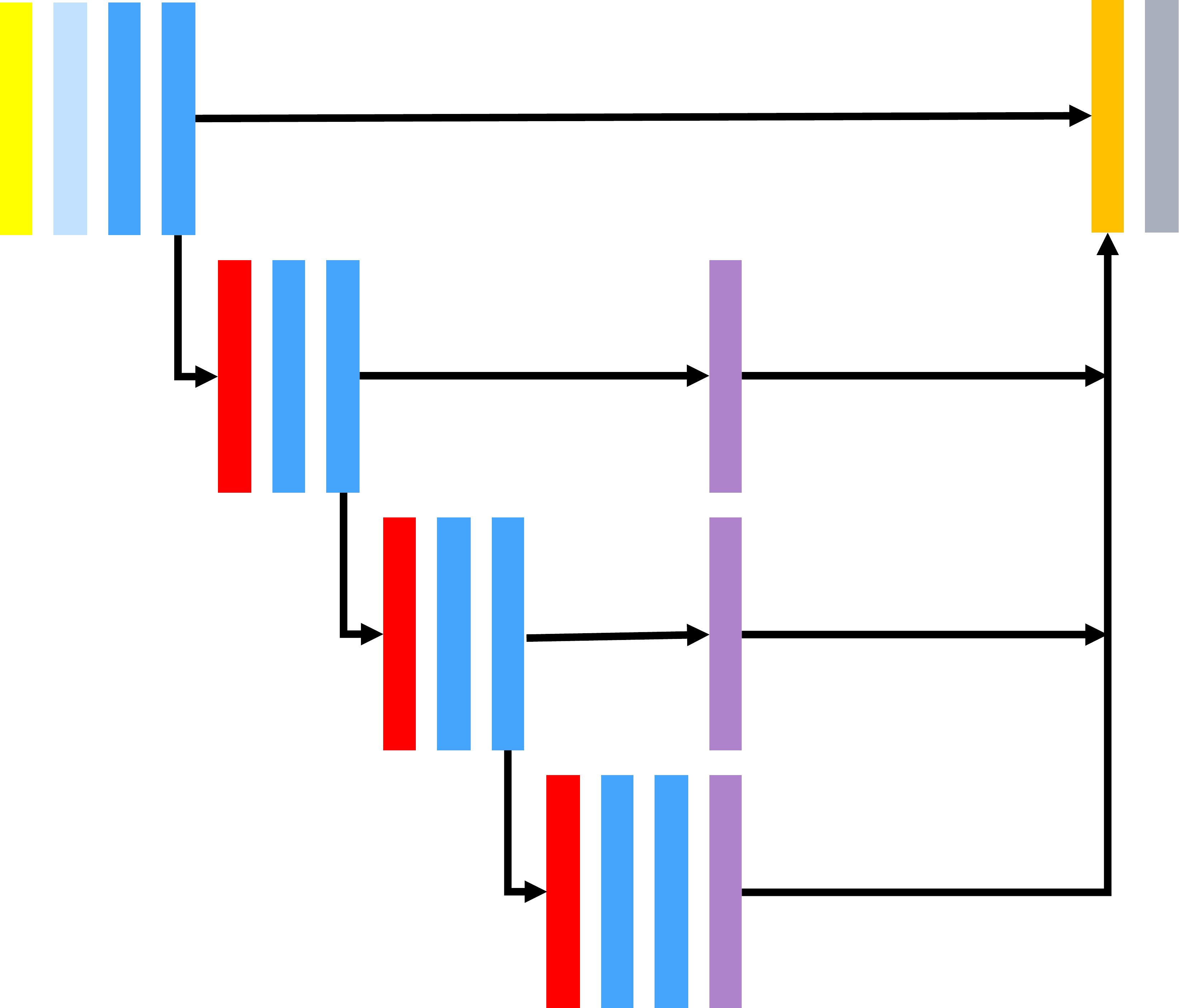}}}
    \qquad
    \subfloat[Submanifold sparse U-Net.]{{\includegraphics[width=0.35\textwidth,height=4cm]{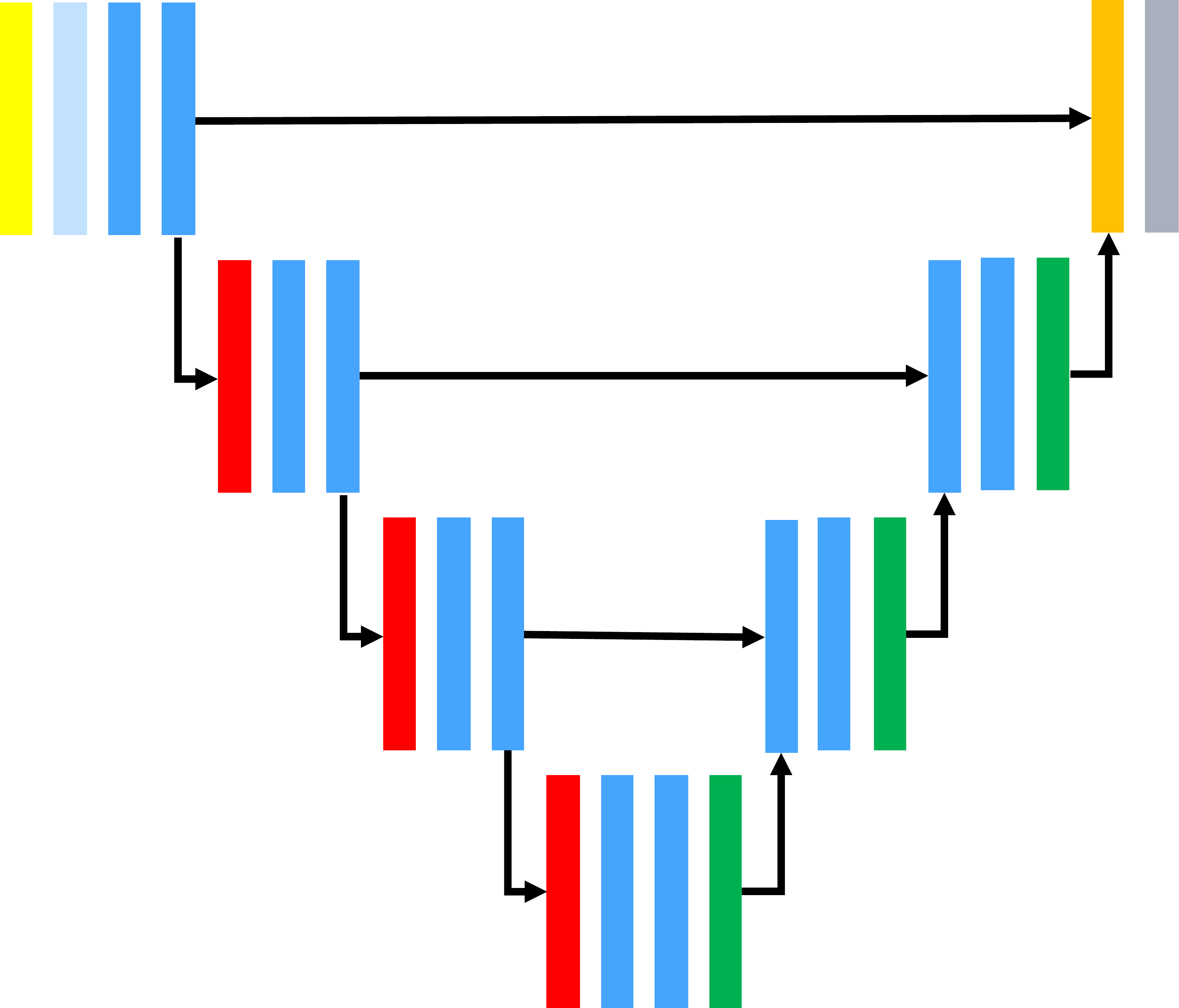}}}
    \qquad
    \subfloat{{\includegraphics[width=0.118\textwidth]{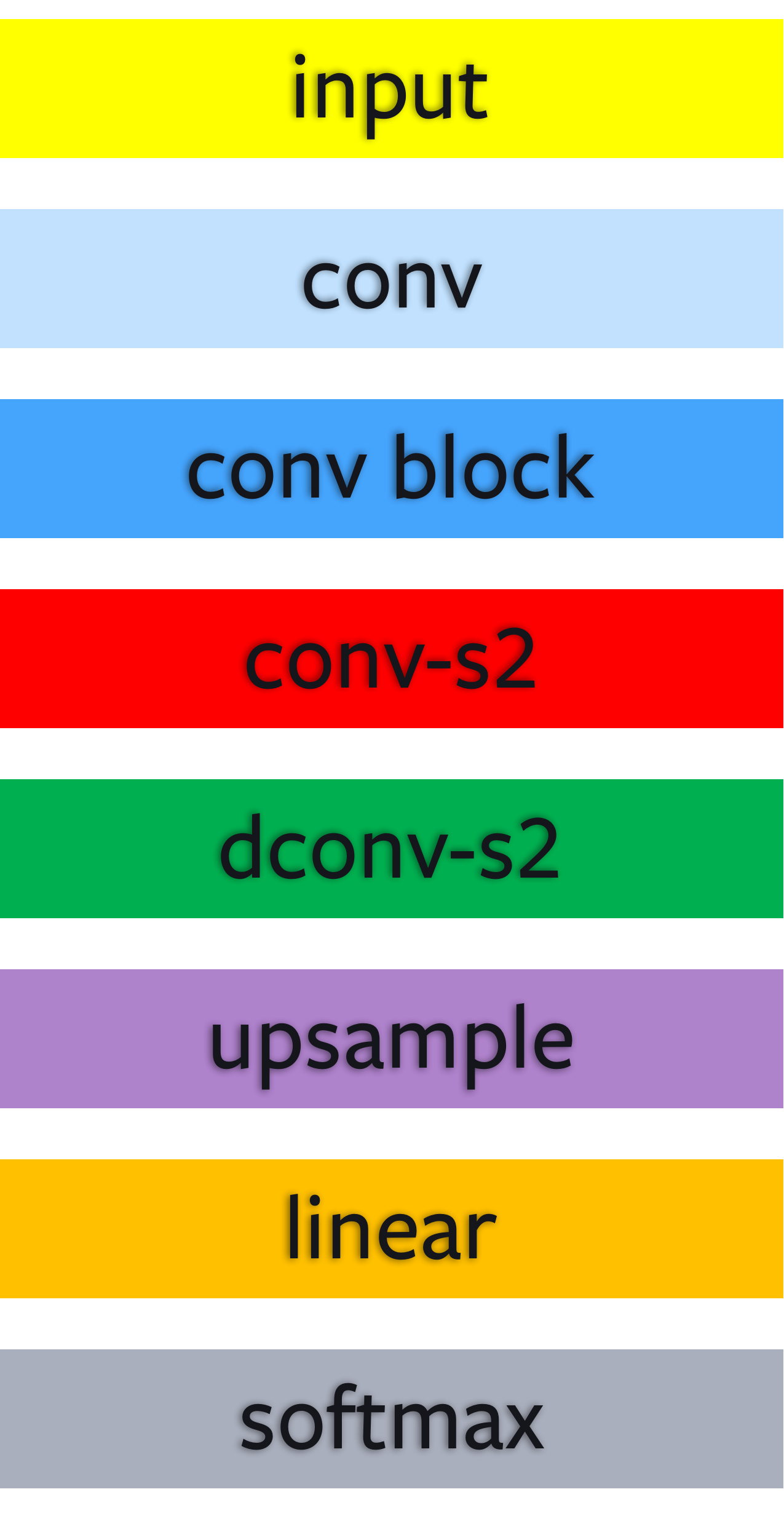}}}
    \caption{Illustrations of our submanifold sparse FCN (a) and U-Net (b) architectures. Dark blue boxes represents one or more ``pre-activated'' SSC$(\cdot,\cdot,3)$ convolutions, which may have residual connections. Red boxes represent size-2, stride-2 downsampling convolutions; green deconvolutions ``invert'' these convolutions. Purple upsampling boxes perform ``nearest-neighbor'' upsampling. The final linear and softmax layers are applied separately on each active input voxel.
    }
    \label{fig:architecture_diagram}
\end{figure*}

We use a sparse voxelized input representation similar to~\cite{graham2015sparse,engelcke2017vote3deep}, and a combination of SSC convolutions and strided SC convolutions to construct versions of the popular FCN \cite{long2015fully} and U-Net \cite{cciccek20163d} architectures. The resulting convolutional network architectures are illustrated in Figure~\ref{fig:architecture_diagram}; see the associated caption for details. We refer to these networks as \emph{submanifold sparse convolutional networks} (SSCNs), because they process low-dimensional data living in a space of higher dimensionality.\footnote{We note that this is a slight abuse of the term ``submanifold''. We emphasize that the data on which these networks are applied may contain multiple connected components, and even a mixture of 1D and 2D objects embedded in 3D space.}

The basic building block for our models are ``pre-activated'' SSC$(\cdot,\cdot,3)$ convolutions. Each convolution is preceded by batch normalization and a ReLU non-linearity. In addition to FCN and U-Nets with standard convolutional layers, we also experiment with variants of these networks that use pre-activated residual blocks \cite{ResNet} that contain two SSC$(\cdot,\cdot,3)$ convolutions. Herein, the residual connections are identity functions: the number of input and output features are equal.
Whenever the networks reduce the spatial scale by a factor of two, we use SC$(\cdot, \cdot,2,2)$ convolutions. Our implementation of FCNs upsamples feature maps to their original resolution rather than performing deconvolutions using residual blocks. This substantially reduces the number of parameters and the number of multiplication-addition operations of the FCN.

%% file: largescale.tex
\section{Experiments}
In this section, we perform experiments with SSCNs on the ShapeNet competition dataset from~\cite{yi2017large}. We compare SSCNs against three strong baseline models in terms of performance and computational cost. Specifically, we consider shape contexts~\cite{belongie2002shape}, dense 3D convolutional networks, and 2D multi-view convolutional networks \cite{conf/iccv/SuMKL15} as baselines. Throughout our experimental evaluation, we focus on the trade-off between segmentation accuracy and computational efficiency measured in FLOPs\footnote{We ignore the FLOPs from the final classification layer.}. In a second set of experiments, we also study SSCN performance on the NYU Depth (v2) dataset~\cite{silberman2012indoor}.

\subsection{Dataset}
The ShapeNet segmentation dataset~\cite{yi2017large} comprises 16 different object categories (plane, chair, hat, \emph{etc.}), each of which is composed of up to 6 different parts. For instance, a ``plane'' is segmented into ``wings'', ``engine'', ``body'', and ``tail''.
Across all object categories, the dataset contains a total of 50 different object part classes. Each object is represented as a 3D point cloud that was obtained by sampling points uniformly from the surface of the underlying CAD model. Each point cloud contains between $2,000$ and $3,000$ points. To increase the size of the validation set, we re-split the training and validation sets using the first bit of the MD5 hash of the point cloud files to obtain a training set with 6,955 examples and a validation set with 7,052 examples. The test set contains 2,874 examples.

In the original dataset, the objects are axis-aligned: for instance, rockets always point along the $z$-axis. To make the problem more challenging, we perform a random 3D translation and rotation on each point cloud before classifying it. The results in Table \ref{tbl:shapeContext} show that removing the alignment, indeed, makes the segmentation task more challenging.

\begin{table}[tb]
\vspace{3pt}
\centering{}
\begin{tabular}{ccccc}
\toprule
\textbf{View type}  &\textbf{IoU accuracy}\\
\midrule
Aligned & 63.5\%\\
Random pose & 47.8\%\\
\bottomrule
\end{tabular}
\caption{Accuracy of segmentation classifiers based on shape-context features on (1) the original ShapeNet dataset and (2) a variant of the dataset in which objects are randomly rotated. The results show that removing the alignment of the ShapeNet objects via random 3D rotations makes the segmentation problem more challenging.\label{tbl:shapeContext}}
\end{table}

To evaluate the accuracy of our models, we adopt the intersection-over-union (IoU) metric of \cite{yi2017large}.
The IoU is computed for each part per object category and averaged over parts and examples for the category to produce a ``per-category IoU''. This way of averaging the IoU scores rewards models that make accurate predictions even for object-parts that are very small: small parts have the same weight in the final accuracy measure as larger parts. The final accuracy measure is obtained by taking a weighted average of the per-category IoUs, using the fraction of training examples per category as weights.

\subsection{Details of Experimental Setup}
\label{sec:details_exp}
In all experiments, the same data pre-processing procedure is used. Specifically, each point cloud is centered and re-scaled to fit into a sphere with diameter $S$; scale $S$ determines the size of the voxelized representation. We use $S\in\{16,32,48\}$ in our experiments. At scale $S=48$, the voxels are approximately 99\% sparse. In experiments with dense convolutional networks, we place the sphere randomly in a grid of size $S$. For submanifold sparse convolutional networks, we place the sphere randomly in a grid of size $4S$.
To voxelize the point cloud, we measure the number of points per voxel and normalize them so that non-empty voxels have a mean density of one.

Networks are trained using the same optimization hyperparameters, unless otherwise noted. We use stochastic gradient descent (SGD) with a momentum of 0.9, Nesterov updates, and $L_2$ weight decay of $10^{-4}$. The initial learning rate is set to $0.1$, and the learning rate is decayed by a factor of $\mathrm{e}^{-0.04}$ after every epoch.
We train all networks for $100$ epochs using a batch size of $16$. We train a single network on all $16$ object categories jointly using a multi-class negative log-likelihood loss function over all $50$ part labels.

For our SSCNs, we experiment with two types of network architectures. The first architecture (C3) operates on a single spatial resolution by stacking SSC$(\cdot,\cdot,3)$ convolutions; we use with $8$, $16$, $32$, or $64$ filters per layer, and $2$, $4$, or $6$ layers. The second architecture type comprises FCNs and U-Nets with three layers of downsampling. These networks have $8$, $16$, $32$, or $64$ filters in the first layer, and double the number of filters each time the data is downsampled. For the convolutional blocks in these networks, we use stacks of $1$, $2$, or $3$ SSC convolutions, or stacks of $1$, $2$, or $3$ residual blocks.

\paragraph{Details on testing.}
At test time, we only compute softmax probabilities for part labels that actually appear in the object that is being segmented, \emph{i.e.}, we assume the models know the category of the object they are segmenting. Softmax probabilities for irrelevant part classes are set to zero (and the distribution over part labels is re-normalized).

For each of the three network types (C3, FCN, and U-Net), we train a range of models with varying sizes, as described above, and monitor their accuracy on the validation set. For each network type, we select the networks that correspond to local maxima on the accuracy \emph{vs.} FLOPs curve, and report test set accuracies for those networks. Akin to multi-crop testing that is commonly used with 2D convolutional networks, we measure test accuracies using ``multi-view'' testing. In particular, we generate $k$ different views of the object with $k \in \{1, \dots, K\}$ by randomly rotating them, and average the model predictions for each point over the $k$ different views of the object.

\subsection{Baselines}
In addition to SSCNs, we consider three baseline models in our experiments: (1) shape contexts~\cite{belongie2002shape}; (2) dense 3D convolutional networks; and (3) 2D multi-view convolutional networks \cite{conf/iccv/SuMKL15}. Details of the four baseline models are described separately below.

\paragraph{Shape contexts.}
Inspired by \cite{belongie2002shape}, we define a voxelized shape context vector. Specifically, we define a \texttt{ShapeContext} layer as a special case of the SSC$(1,27,3,1)$ submanifold convolution operator: we set the weight matrix of the operator to be a $27 \times 27$ identity matrix so that it accumulates the voxel intensities in its $3^3$ neighborhood.

We scale the data using average pooling with sizes 2, 4, 8 and 16 to create four additional views. Combined this gives each voxel a 135-dimensional feature vector. The feature vector is then fed into a non-convolutional multi-layer perceptron (MLP) with two hidden layers, followed by a 50-class softmax classifier. The MLPs have 32, 64, 128, 256, or 512 units per layer. At test time, we use multi-view testing with $K=3$.

\begin{figure*}[t]
    \centering
    \subfloat[Comparison with baseline methods.]{{\includegraphics[width=0.33\textwidth]{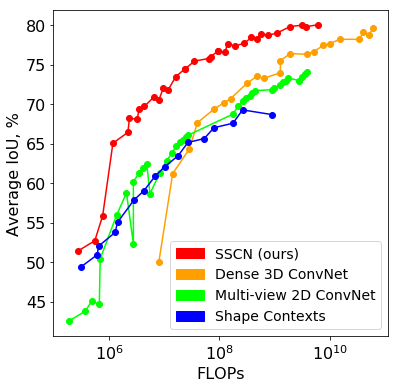}}}
    \subfloat[Comparison between architectures (see \ref{sec:details_exp}).]{{\includegraphics[width=0.33\textwidth]{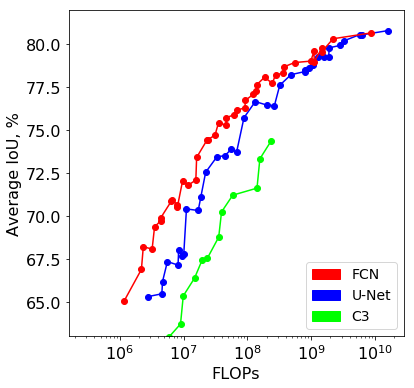}}}
    \subfloat[SSCN with different scales, $S$.]{{\includegraphics[width=0.33\textwidth]{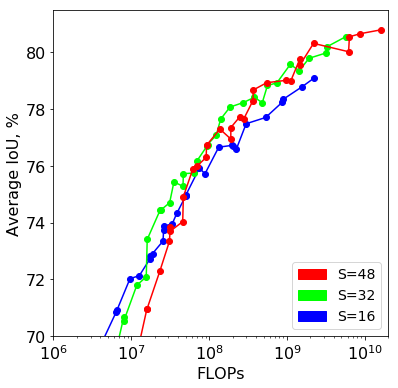}}}
    \caption{Average interaction-over-union (IoU) on the test set of SSCNs trained for 3D semantic segmentation on the ShapeNet competition data set (higher is better).}
    \label{fig:results}
\end{figure*}

\paragraph{Dense 3D convolutional networks.}
For dense 3D convolutional networks, we simply considered dense versions of the SSCN networks. Due to computational constraints, we restricted the FCN and U-Net convolutional blocks to a single C3-layer. We trained some of the models with a reduced learning rate due to numerical instabilities we observed during training. Again, we use $K=3$ multi-view testing.

\paragraph{Convolutional networks on multi-view 2D projections.}
This baseline model discards the inherent 3D structure of the data by projecting the point cloud into a two-dimensional view by assuming infinite focal length, applying a 2D convolutional network on this projection, and averaging the predictions over multiple views. An immediate advantage of this approach is that well-studied models from 2D vision can be used out-of-the-box without further adaptations. Moreover, the computational cost scales with the surface area, rather than the volume of the point cloud.

In our implementation of this approach, we first convert the point clouds into a 3D grid of size $S^3$ as we did for the previous baseline. We then project to a plane of size $S^2$, i.e. a face of the cube, with two feature channels. One feature channel is the first visible, non-zero voxel along the corresponding column. The second channel is the distance to the visible voxel, normalized to the range $[0,2]$; this is analogous to the depth channel of an RGB-D image. Our network architectures are two-dimensional versions of the dense 3D convolutional networks described above.

During training, a random projection of the point cloud is passed into the model. Points in the point cloud that fall into the same voxel are given the same prediction. Some voxels are occluded by others---the network receives no information on the occluded voxels.
We modify the multi-view testing procedure to take into account the occlusion of voxels.
Similar to before, predictions are performed using a weighted sum over $k$ random projections. We found that 2D networks require more views to obtain high accuracy and therefore use $K=10$. Voxels that are observed in the 2D projection are given a weight of $1$. The weight of occluded voxels decays exponentially with the distance to the voxel that occludes them.

\subsection{Results}
In Figure \ref{fig:results}, we report the average IoU of a range of differently sized variants of (1) the three baseline models and (2) submanifold C3, FCN and U-Nets on the ShapeNet test set. The average IoU is shown as a function of the number of multiplication-addition operations (FLOPs) required by the models for computing the predictions. Note that these results are not directly comparable with those in \cite{yi2017large} because we are testing the models in the more challenging ``random-pose'' setting.

\paragraph{SSCNs vs. baselines.} Figure \ref{fig:results}(a) compares SSCNs with the three baselines.\footnote{The number of FLOPs reported for shape contexts may be slightly misleading: the computational costs of calculating shape context features is not reflected in the number of FLOPs, as it involves integer arithmetic.}
The results show that shape context features, multi-view 2D ConvNets, and dense 3D ConvNets perform roughly on par in terms of accuracy per FLOP. SSCN networks outperform all baseline models by a substantial margin. For instance, at $10^8$ FLOPs, the average IoU of SSCNs is $6$-$8\%$ higher than that of the baselines. Importantly, our results show that restricting information to travel along submanifolds in the data does not hamper the performance of SSCNs, whilst it does lead to considerable computational and memory savings that can be exploited to train larger models with better accuracies.

\paragraph{Ablation.} In Figure \ref{fig:results}(b), we compare the three SSCN architectures presented in Section~\ref{sec:details_exp}. We observe that SSCNs involving downsampling and upsampling operations (FCNs and U-Nets) outperform SSCNs operating on a single spatial resolution and we conjecture that this is due to the increased receptive field obtained by downsampling.

Figure \ref{fig:results}(c) shows the performance of SSCNs at three different scales $S$ (using all three architectures; C3, FCN, and U-Net). We observe that the performance of SSCNs is similar for different values of $S$, particularly for a low number of FLOPs. At a higher number of FLOPs, the models operating at a larger scale perform slightly better.

\subsection{Results on Competition Data}\label{subsec:aligned}
To compare SSCNs with the entries to the competition in \cite{yi2017large}, we also trained a FCN SSCN on the aligned point clouds. In this experiment, we performed data augmentation using random affine transforms. We set $S \!=\! 24$ and use $64$ filters in the input layer, three levels of downsampling, and two residual blocks per spatial resolution. The results of 10-view testing are compared with the competition entries in Table~\ref{tab:seg_results}. With a test error of 85.98\%, our network outperforms other methods by $\geq\! 0.49\%$ IoU.

%% file: nyu.tex
\subsection{Semantic Segmentation of Scenes}
\begin{figure}[t]
\includegraphics[angle=270,width=0.49\columnwidth]{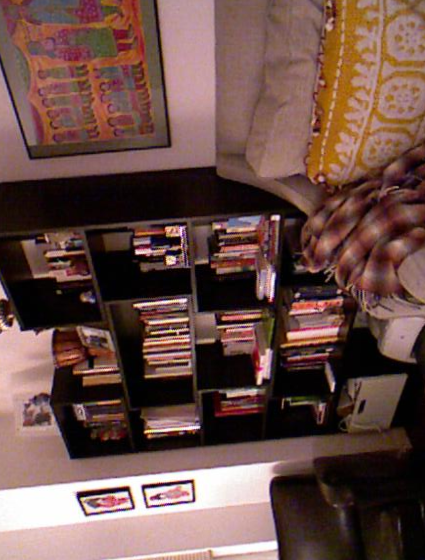}\ \includegraphics[angle=270,width=0.49\columnwidth]{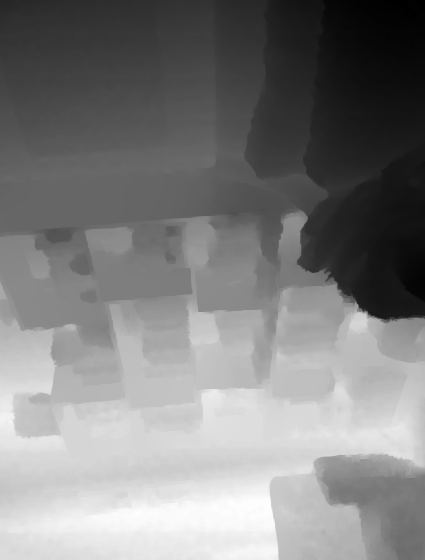}\\
\includegraphics[angle=270,width=0.49\columnwidth]{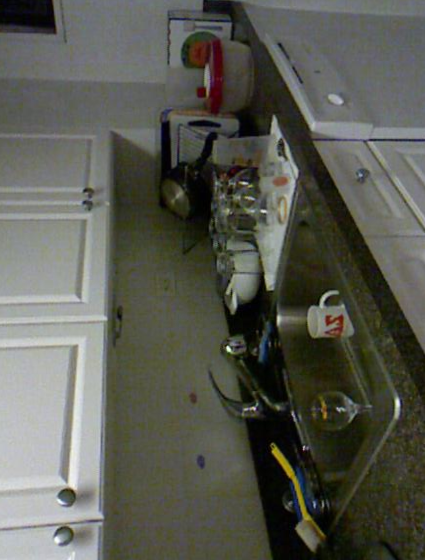}\ \includegraphics[angle=270,width=0.49\columnwidth]{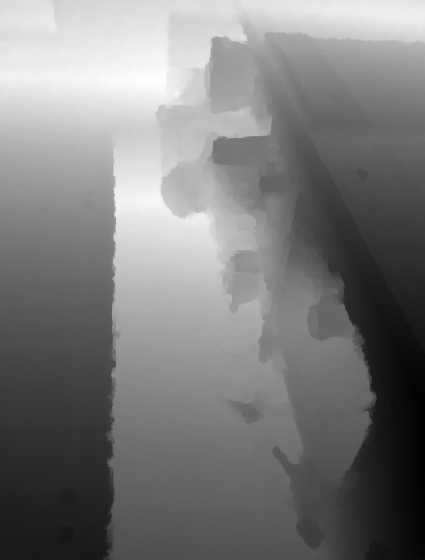}\\
\caption{Two examples of RGB-D images from the NYU Depth dataset (v2)~\cite{silberman2012indoor}.}
\label{fig:nyu}
\end{figure}

We also performed experiments on the NYU Depth dataset (v2)~\cite{silberman2012indoor} for semantic segmentation of scenes rather than objects. The dataset contains $1,449$ RGB-D images, which are semantically segmented into 894 different classes. Figure~\ref{fig:nyu} shows two pairs of an RGB image and the associated depth map from the dataset. Following \cite{conf/cvpr/GuptaAM13,long2015fully}, we crop the images and reduce the number of classes to $40$. To assess the performance of our models, we measure their pixel-wise classification accuracy. We compare our models to a 2D FCN \cite{long2015fully}.

We perform experiments with two differently sized SSCN-FCN networks.
Network A has 16 filters in the input layer, and one SSC$(\cdot,\cdot,3)$ convolution per level.
Network B has 24 filters in the input layer, and two SSC$(\cdot,\cdot,3)$ convolutions per level.
Both networks use eight levels of downsampling; we increase the number of filters in the networks when downsampling, adding 16 (A) or 24 (B) features with each reduction of scale.

We use the depth information to convert the RGB-D images into a 3D point cloud. Each point in the cloud has the three (RGB) features that were normalized to the range $[-1,1]$, and a fourth indicator features that is set to $1$ for each point in the point cloud. During training, we perform data augmentation by applying random affine transformations to the point cloud. Before voxelizing the point cloud, we downscale by a factor of two, and place the points into the model's receptive field. We form voxels by averaging the feature vectors of the points corresponding to the voxel. At test time, we perform multi-view testing with $k =1, 4$.

The results of our experiments on the NYU Depth dataset (v2) are presented in Table \ref{tbl:nyu}. The results in the table show that SSCNs outperform 2D FCN in terms of pixel accuracy by up to $7\%$, whilst also substantially reducing the computational costs of the model.

To verify that SSCN-FCN A actually uses depth information, we repeat the experiment whilst setting all the depth values to zero; this prevents the SSCN from exploiting depth information. We observe: (1) a reduction of FLOPs by 60\%, as there are fewer active voxels; and (2) a drop in accuracy from 64.1\% to 50.8\%, which demonstrates that SSCNs do use 3D structure when performing segmentation.

\begin{table}[t]
\centering{}
\begin{tabular}{lccccc}
    \toprule
    \textbf{Network}        &\textbf{$k$} &~& \textbf{Accuracy}  & \textbf{FLOPs}  &  \textbf{Memory}\\
    \midrule
    2D FCN \cite{long2015fully}  &1 && 61.5\% & 28.50G & 135.7M \\
    \midrule
    \multirow{2}{*}{SSCN-FCN A} &1 && 64.1\% &  1.09G &  5.2M  \\
     &4 && 66.9\% &  4.36G & 20.7M  \\
    \multirow{2}{*}{SSCN-FCN B}  &1 && 66.4\% &  4.50G & 11.6M  \\
    &4 && 68.5\% & 17.90G & 46.4M  \\
    \bottomrule
\end{tabular}
\caption{Semantic segmentation performance of five different convolutional networks on the NYU Depth test set (v2) on 40 classes. We report, the pixel-wise classification accuracy, the computational costs (in FLOPs), and the memory requirements (c.f. Table~\ref{tbl:flops}).}\label{tbl:nyu}
\end{table}

%% file: conclusions.tex
\section{Conclusions}
In this paper, we presented submanifold sparse convolutional networks (SSCNs) for the efficient processing of high-dimensional, sparse input data. We demonstrated the efficacy of SSCNs in a series of experiments on semantic segmentation of three-dimensional point clouds. Specifically, our SSCN networks outperform a range of state-of-the-art approaches for this problem, both when identifying parts within an object and when recognizing objects in a larger scene. Moreover, SSCNs are computationally efficient compared to alternative approaches.